\documentclass{article} 
\usepackage{nips14submit_09,times}
\usepackage{epsfig}
\usepackage{graphicx}
\usepackage{amsmath}
\usepackage{amssymb}
\usepackage{subfig}
\usepackage{psfrag}
\usepackage{float}
\usepackage{mathtools}
\usepackage{algorithm}
\usepackage[noend]{algorithmic}
\DeclareMathOperator{\det1}{det}

\usepackage[pagebackref=true,breaklinks=true,letterpaper=true,colorlinks,bookmarks=false]{hyperref}

\title{ Multi-Object Classification and Unsupervised Scene Understanding Using Deep Learning Features and
Latent Tree Probabilistic Models} 
\author{Tejaswi Nimmagadda\\
University of California, Irvine\\
\texttt{nimmagas@uci.edu}
\And
Anima Anandkumar\\
University of California, Irvine\\
\texttt{a.anandkumar@uci.edu}
}

\begin{document}

\maketitle

\begin{abstract}

Deep learning has shown state-of-art classification  performance on datasets such as ImageNet, which  contain a single object in each image. However, multi-object classification  is  far more challenging.   We present a unified framework which leverages the strengths of multiple machine learning methods, viz deep learning, probabilistic models and kernel methods to obtain state-of-art performance on Microsoft COCO, consisting of non-iconic images. We  incorporate contextual information  in natural images through a conditional latent tree probabilistic model (CLTM), where the object co-occurrences are conditioned on the    extracted fc7 features from  pre-trained Imagenet  CNN as input.   We learn the CLTM tree structure  using conditional pairwise probabilities for object co-occurrences, estimated through  kernel methods, and we learn its node and edge potentials by training a new 3-layer neural network, which takes fc7 features as input. Object classification is carried out via inference on the learnt conditional tree model, and  we obtain  significant  gain in precision-recall and F-measures on MS-COCO, especially for  difficult object categories. Moreover, the latent variables in the CLTM capture scene information: the images with top activations for a latent node have common  themes such as being a grasslands or a food scene,  and on on. In addition, we show that a simple  k-means clustering of the inferred latent nodes alone significantly improves scene classification performance on   the MIT-Indoor dataset, without the need for any retraining, and without using  scene labels during training. Thus, we present a unified framework for multi-object classification and unsupervised scene understanding.
\end{abstract}


\section{Introduction}

Deep learning has revolutionized performance on a variety of computer vision tasks such as object classification and localization, scene parsing, human pose estimation, and so on. Yet, most deep learning works focus on simple classifiers at the output, and train on   datasets such as ImageNet which consist of single object categories. On the other hand, multi-object classification is a far more challenging problem.

Currently many frameworks for multi-object classification use simple approaches:  the multi-class setting, which predicts one category out of a set of mutually exclusive categories (e.g. ILSVRC~\cite{ILSVRC15}), or binary classification, which makes binary decisions for each label independently (e.g. PASCAL VOC~\cite{pascal}). Both models, however, do not capture the complexity of  labels in natural images. The labels are not mutually exclusive, as assumed in the multi-class setting.  Independent binary classifiers, on the other hand, ignore the relationships between labels and miss the   opportunity to transfer and share knowledge among different label categories during learning. More sophisticated classification techniques based on
structured prediction are being explored, but in general, they  are  computationally more expensive and not scalable to large datasets (see related works for a discussion).

In this paper, we propose an efficient multi-object classification framework by   incorporating contextual information in images.  The context in natural images captures relationships between various object categories, such as co-occurrence of objects within a scene or relative positions of objects with respect to a background scene. Incorporating such contextual information   can vastly improve  detection performance, eliminate false positives, and provide a coherent scene interpretation.


We present an efficient and a unified approach to learn contextual information through probabilistic latent variable models, and combine it with pre-trained deep learning features  to obtain state-of-art   multi-object classification system. It is known that deep learning produces transferable features, which can be used to learn new tasks, which differ from tasks on which the neural networks were trained~\cite{transferable,oquab2014learning}. Here,  we demonstrate that    the transferability of pre-trained deep learning features  can be further enhanced by capturing the contextual information in images.


We model the contextual   dependencies using a conditional latent  tree model (CLTM), where we condition on the pre-trained deep learning features as input. This allows us to incorporate the joint effects of both the pre-trained features and the context for object classification.
Note that a hierarchical tree structure is natural for capturing the groupings of various object categories in images; the latent or hidden variables capture the ``group'' labels   of objects. Unlike previous works, we do not impose a fixed tree structure, or even a fixed number of latent variables, but learn a flexible structure efficiently from data. Moreover, since we make these ``group'' variables latent, there is no need to have access to  group labels during  training, and we  learn the object groups or scene categories in a unsupervised manner. Thus, in addition to efficient multi-object classification, we also learn latent variables that capture semantic information about the scene in a unsupervised manner.

\subsection{Summary of Results}

We propose a unified framework for multi-object classification and scene understanding that combines the strengths of multiple machine learning techniques, viz deep learning, probabilistic models, and kernel methods. We demonstrate significant improvement over state-of-art deep learning methods, especially on challenging objects. We learn a conditional latent tree model, where we condition on pre-trained deep learning features. We employ kernel methods to learn the structure of the hierarchical tree model, and we train a new smaller neural network to learn the node and edge potentials of the model. Multi-object classification is carried out via  inference on the tree. All these steps are efficient and scalable to large datasets with a large number of object categories.



We extract features using pre-trained ImageNet CNN~\cite{imagenet} from Caffe~\cite{caffe}, and use it as input to the conditional latent tree model (CLTM), a type of conditional random field (CRF).   The tree dependency structure for this model is recovered using distance based methods~\cite{LTM}, which requires pairwise conditional probabilities of object co-occurrences, conditioned on the input features. We employ the kernel conditional embedding framework~\cite{song} to compute these pairwise measures. Using a feed-forward neural network, we   train the above energy based model; the outputs of this neural network yield the node and edge potentials of the CLTM. We test performance of multi-object classification on a non-iconic image set Microsoft COCO~\cite{MSCOCO} and we test its unsupervised scene learning capabilities on the MIT Indoor dataset~\cite{objectbank}.


We  recover a natural coherent tree structure on the MS COCO data-set, using training images, each of which contain only few object categories. For instance, objects (e.g. table, chair and couch) that appear in a given scene (living room) are grouped together.
Using our approach, precision-recall performance and F-measures are significantly improved compared to the baseline of a 3-layer neural network with independent binary classifiers, which also takes in fc7 features as input. We see across the board improvement for all object categories  over the entire precision-recall curve. The
overall   relative gain in F-measure for our method is 7\%.   For difficult objects like couch, frisbee, cup, bowl, remote, fork, and wine-glass,  the F-measure relative gain is 41\%,  48\%, 50\%, 53\%,  113\%,  122\%, and 171\% respectively.   Thus, we combine   pre-trained deep learning features and the learnt contextual model to obtain state-of-art multi-object classification performance.



We also demonstrate how   latent nodes can be used for unsupervised scene understanding, without using any scene labels during training. We observe that latent nodes capture high-level semantic information   common to images, based on the   neighborhoods of object categories in the latent tree.  When we consider the top  images with largest activations of node potential for a given latent node, we find diverse images with different objects, but with a unifying  common theme. For instance, for one of the latent variables, the top images capture  a grassland scene but with  different animals in different images.  Similarly,  the latent variable representing an outdoor scene contains diverse images with traffic, beaches, and buildings. As another example, the latent variable representing the food scene shows foods of various different kinds. Thus, we present a flexible framework for capturing thematic information in images in a unsupervised manner.

We also  quantitatively show that the latent variables yield efficient scene classification performance on the  MIT-Indoor dataset, without any re-training, and without using any scene labels during   training. We use the marginal probabilities  of the latent variables in our model on   test images, and perform $k$-means clustering. For validation, we match these clusters to ground truth scene categories using maximum weight matching~\cite{ahuja1989network}. We obtain 20\% improvement in misclassification rate of the scenes, compared to the neural network baseline. Note that we assume that the scene labels are not present during training for both our method, and for the neural network baseline. Thus, we demonstrate that our model is capable of capturing rich semantic information about the scenes, without using any scene labels during the training process.

Thus, we present a carefully engineering unified framework for multi-object classification that combines the strengths of diverse machine learning techniques. While general non-parametric methods are computationally expensive, and not scalable to large datasets, we employ kernel methods only to estimate pairwise conditional probabilities, which can be carried out efficiently using randomized matrix techniques~\cite{nystorm}. Our tree structure estimation is scalable to large datasets using recent advances in parallel techniques for structure estimation~\cite{scalableLTM}. Instead of training a large neural network from scratch, we train a smaller one, and we use a energy-based model at its output to  obtain the node and edge potentials of the latent tree model. Finally, at test time, we have ``lightning'' fast inference using message passing on the tree model. Thus, we present an efficient and  a scalable framework for handling large image datasets with a large number of object categories.




 \subsection{Related Work}


Correlations between labels have been explored for detecting multiple object categories before.~\cite{myungjin,outofcontext} learn contextual relations between co-occurring objects using a tree structure graphical model   to capture dependencies among different objects. In this model, they incorporate  dependencies between object categories, and outputs of local detectors into one probabilistic framework.   However, using simple pre-trained object detectors   are typically noisy and lead to performance degradation. In contrast, we employ pre-trained deep learning features as input, and consider a conditional model for context, given the features. This allows us to incorporate both deep learning features and context into our framework.


In many settings, the hierarchical structure representing the contextual relations between different objects is fixed and is based on semantic similarity~\cite{grauman2011learning}, or may rely on text, in addition to image information~\cite{li2010building}. In contrast, we learn the tree structure from data efficiently, and thus, the framework can be adapted to settings where such a tree may not be available, and even if available, may not give the best classification performance for multi-object classification.

Using pre-trained ImageNet features for other computer vision tasks has been popular in a number of works recently, e.g.~\cite{transferable,girshick2014rich,oquab2014learning}.~\cite{girshick2014rich} term this as {\em supervised pre-training} and employ them to train regional convolutional neural networks (R-CNN) for object localization. We note that our framework can be extended to
localization and we plan to pursue it in future.  While~\cite{girshick2014rich} employ independent SVM classifiers for each class, we believe that incorporating our probabilistic framework for multi-object localization can significantly improve performance.
Recently,~\cite{zhang2015improving} propose improving object detection using Bayesian optimization for fine grained search and a structured loss function that aims at both classification and localization. We believe that incorporating probabilistic contextual models can further improve performance in these settings.

Recent papers also incorporate deep learning for scene classification.~\cite{zhou2014learning,zhou2014object} introduce the places dataset and use CNNs for scene classification. In this framework, scene labels are available during training, while we do not assume access to these labels during our training process. We demonstrate how introducing latent variables can automatically capture semantic information about the scenes, without the need for labeled data.


Scene understanding is a very rich and an active area of computer vision and consists of a variety of tasks such as object localization, pixel labeling, segmentation and so on, in addition to classification tasks.~\cite{li2009towards} propose a hierarchical generative model that performs multiple tasks in a coherent manner. \cite{li2011theta} also consider the use of context by taking into account the spatial location of the regions of interest.
While there is a large body of such works which use contextual information (see for instance~\cite{li2011theta}), they mostly  do not incorporate latent variables in their modeling.  In future, we plan to extend our framework for these various scene understanding tasks and expect significant improvement over existing methodologies.




There have been some recent attempts to combine neural networks with probabilistic models. For example,~\cite{ammar2014conditional} propose to combine CRF and auto-encoder frameworks for unsupervised learning. Markov random fields are employed for  pose estimation to encode the spatial relationships between joint locations in~\cite{tompson2014joint}.~\cite{chen2014learning} propose a joint framework for deep learning and probabilistic models. They learn deep features which take into account dependencies between output variables. While they train a 8-layer deep network from scratch to learn the potential functions of a MRF, we exhibit how a simpler network can be used if we employ pre-trained features as an input to the conditional model. Moreover, we incorporate latent variables that allow us to use a simple tree model, leading to faster training and inference.  Finally, while many works have used MS-COCO for captioning and joint image-text related tasks~\cite{karpathy2014deep,vinyals2014show}, there have been no attempts to improve multi-object classification  over standard deep learning techniques, using images alone on MS-COCO and not the text data, to the best of our knowledge.

The rest of this paper is organized as follows. Section~\ref{sec:2}
presents overview of the model. Section \ref{sec:3} presents structure learning method using input distribution of fc7 features. In Section~\ref{sec:4}, we discuss how we train CLTM using neural networks. In Section~\ref{sec:5}, we evaluate the proposed model on MS COCO dataset and discuss the results.  Finally, Section~\ref{sec:conclusion} concludes the paper.

\section{Overview of The Model And Algorithm}\label{sec:2}

\begin{algorithm}[t]
 \caption{Overview of the Framework}
 \label{algo:overview}
 \begin{algorithmic}[1]
 \REQUIRE Labeled image-set $\mathcal{I} = \{(I^{1},y^{1}),\cdots,(I^{n},y^{n})\}$

   \STATE {$\{x^{1},x^{2},\cdots,x^{n}\} \leftarrow $ ExtractFc7Features($\mathcal{I}$)}
     \STATE{
  Estimate conditional distance matrix : \\ D $\leftarrow $ CondDistanceMatrix($\{(x^{1},y^{1}),\cdots,(x^{n},y^{n})\}$) \\
   using kernel methods.
}
   \STATE{ Extract tree structure using ~\cite{LTM}\\
    $\mathcal{T} \leftarrow$ CLRG(D)}

\STATE{Training a NN with randomly initialized weights W:}

 	\REPEAT
	\STATE{randomly select a mini-batch $M$.} \STATE {compute negative marginalized log-likelihood loss: Eqn.\eqref{loss}\\ $\mathcal{L} \leftarrow$  Loss(W,$\mathcal{T},M$)}
	\STATE{W  $\leftarrow$ BackpropogateGradient($\mathcal{L}$)
	} \UNTIL{convergence}
	
 		\STATE Given a test image $T$: $x^{t} \leftarrow$  ExtractFc7Features($T$)
	 \STATE  Potentials $\leftarrow$ FeedForward(W,$x^{t}$)
	 \STATE   Prediction: $y \leftarrow {\arg\min}_{ Y } Energy(Y,Potentials)$

 \end{algorithmic}
 \end{algorithm}

We consider pre-trained ImageNet \cite{imagenet} as  a fixed feature extractor by considering the fc7 layer (4096-D vector) as the feature vector for a given input image.  We denote this extracted feature  as $x^{i}$  for $i^{th}$ image.  It is also demonstrated in \cite{transferable} that such feature vectors can be effectively used for different tasks with different labels. The goal here is to learn models which can label an image to multiple-object categories present in a given image. Our model predicts a structured output  $y \in \{0,1\}^{L}$.  To achieve this goal ,we use a dependency structure that relates different object labels. Such dependency structure should able to capture pair-wise probabilities of object labels conditioned on input features. We model this dependency structure using a latent tree. Firstly, these type of structures allow for more complex structures of dependence compared to a fully observed tree. Secondly, inference on it is tractable.

We estimate probabilities of object co-occurrences conditioned on input fc7 features. We then use distance-based algorithm to recover the structure using estimated distance matrix.  Once we recover the structure, we model the distribution of observed labels and latent nodes for a given input covariates as a discriminative model. We use conditional latent Tree Model, a class of CRF that belongs to exponential family of distributions to model distribution of output variables given an input. Instead of restricting the potentials(factors) to linear functions of covariates, we generalize potentials as functions represented by outputs of a neural network. For a given architecture of neural network which takes $X$ as input, we learn weights $W$ by  backpropogating the gradient of marginalized log-likelihood of output binary variables. Once we train the given neural network, we consider the outputs of neural network as  potentials for estimating marginal node beliefs conditioned on input covariates $X$. Our model also results in MAP configuration for a given input covariates $X$. Algo.\ref{algo:overview} gives overview of our framework.

\begin{figure}
\begin{center}
\psfrag{Input Layer}[i]{}
\psfrag{Input Layer - fc7 features}[i]{ \tiny{Input Layer}}
\psfrag{Hidden layer}[l]{ \tiny{Hidden Layer}}
\psfrag{Output Layer}[l]{ \tiny{Output Layer}}
\psfrag{node potential for node h1}[l]{ \tiny{node potential - h1}}
\psfrag{node potential for y3}[l]{ \tiny{node potential - y3}}
\psfrag{latent node}[l]{ \tiny{latent node}}
\psfrag{observed nodes}[l]{ \tiny{observed node}}
\includegraphics[width=.5\linewidth]{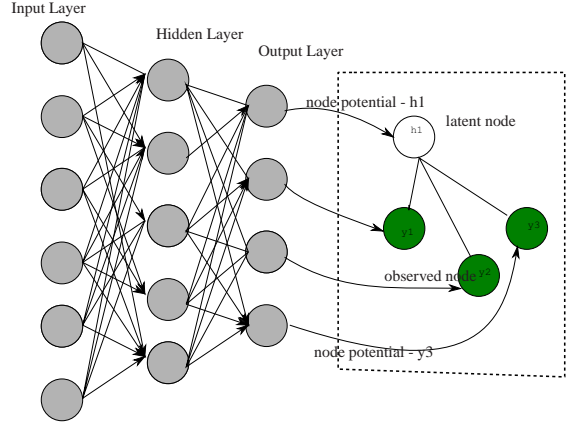}
\end{center}
\vspace{-0.15in}

\caption{ Our Model takes input as fc7 features and generates node potentials at the output layer of a given neural network. Using these node potentials, our model outputs MAP configuration and marginal probabilities of observed and latent nodes}
\label{neuralnet}
\end{figure}

Use of non-parametric methods for end-end tasks on large datasets is computationally expensive. So, we restrict using kernel methods to only evaluate  pairwise conditional probabilities, and here, we can use randomized matrix methods to efficiently scale the computations ~\cite{nystorm}. The tree structure is estimated through CL grouping algorithm from ~\cite{LTM}. Although the method in ~\cite{LTM} is serial, we note that recently there have been parallel versions of this method in ~\cite{scalableLTM}.  Finally, we train neural networks to output node and edge potentials for CLTM. Finally, detection is carried out via inference on the tree model through message passing algorithms. Thus, we have an efficient procedure for multi-object detection in images.

\section{Conditional Latent Tree Model} \label{sec:3}

We denote given labeled training set as $D =\{(x^{1},y^{1}),\cdots,(x^{n},y^{n})\}$  and  $ x^{i} \in  \mathbb{R}^{4096} , y^{i} \in  \{0,1\}^L$ $\forall$  i  $\in  (1,2,\cdots,n)$. We denote extracted tree by $ \mathcal{T} = (\mathcal{Z}, \mathcal{E})$ where $\mathcal{Z}$ indicates the set of observed and latent nodes and $\mathcal{E}$ denotes edge set.  Once we recover the structure, we use conditional latent tree model to model $P( \mathcal{Z}|X)$. Conditioned on input $X$, we model distribution of $\mathcal{Z}$  using in the below Eqn.
\begin{equation*}
P(\mathcal{Z} | X)  = \exp\left(-\sum_{k \in \mathcal{Z}}\phi_{k}(X, \theta)z_{k}+    \sum_{(k,t) \in \mathcal{E}}\phi_{(k,t)}(X,\theta)z_{k}z_{t}  - \mathcal{A}(\theta,X)\right)
\end{equation*}

where $\mathcal{A}(X, \theta)$ is the term that normalizes the distribution, also known as the log partition function. $\phi_{k}(X, \theta)$ and  $\phi_{(k,t)}(X, \theta)$
indicate the node and edge potentials of the exponential family distribution, respectively. Instead of restricting the potentials to linear functions of covariates, we generalize potentials as functions represented by outputs of a neural network. Sec.\ref{sec:4} explains how we learn the weights of such a neural network.

We learn the dependency structure among object labels from a set of fully labeled images. Traditional distance-based methods use only empirical co-occurrences of objects to learn the structure. Learning a structure that involves strong pair-wise relations among objects requires  training images to contain many instances of different object categories.
In this section, we propose a new structure recovery method without the need of such training sets. This method involves both empirical co-occurrences and the distribution of fc7 features to calculate distances between labels.

Since there are very few positive sample images with multiple object-categories, training just based on co-occurrence is not sufficient to recover a coherent tree structure. We  leverage on extracted features to estimate moments by conditioning on them. We propose a new method to calculate the distance matrix by using a RKHS framework to estimate moments. The estimated distance matrix is then used by distance-based methods for structure recovery \cite{LTM}.


\subsubsection*{Kernel Embedding of Conditional Distribution}
The kernel conditional embedding framework, described in \cite{song} gives us methods for modeling conditional and joint distributions.
These methods are effective in high-dimensional settings with multi-modal components such as the current setting .

In the general setting, given transformations $\phi(X)$ and $\Psi(Y)$ on X,Y to the RKHS using kernel functions $K(x,.)$, $K^{'}(y,.)$, the above framework provides us with the following empirical operators to embed joint distributions into the reproducing kernel Hilbert space (RKHS). Define
\begin{align*}
   \hat{\mathcal{C}}_{XX} &= \frac{1}{N} \sum_{n=1}^N\phi(x^{n})\otimes \phi(x^{n}) \\
  \hat{\mathcal{C}}_{XY} &= \frac{1}{N}  \sum_{n=1}^N\phi(x^{n})\otimes \Psi(y^{n}),
\end{align*} and $\hat{\mathcal{C}}_{Y|X} := \hat{\mathcal{C}}_{YX}\hat{C}_{XX}^{-1} $.
We have following results that can be used to evaluate $\hat{\mathbb{E}}_{Y_{i}Y_{j}|X}[y_{i} \otimes y_{j}|x]$ for a given data-set.
  \begin{equation}
 \Psi(y)^{\top}\hat{\mathcal{C}}_{Y|X} \phi(x) = \Psi (y)^{\top} \Psi_{Y} ( K_{XX} + \lambda N I)^{-1} \phi _{X}^{\top} \phi(x)
\end{equation} We employ Gaussian RBF kernels and use the estimated conditional pairwise probabilities for learning the latent tree structure.

 \subsection{Learning Latent Tree Structure}

\begin{algorithm}[h]
 \caption{CondDistanceMatrix}
 \label{algo:distance}
 \begin{algorithmic}[1]
 \REQUIRE Input data-set $D =\{(x^{1},y^{1}),\cdots,(x^{n},y^{n})\}$

  \STATE  Compute Gram matrix  $K_{n\times n}$ using hyper-parameter $ \gamma$
 	\FOR{$ i = 1$ TO $i =n$}
	
 	\STATE $G = (K + \lambda I)^{-1} \times K(:,i)$.
	\FOR{ all pairs (k,t) where  k,t  $\in$ $(1,2,\cdots,L)$}
	\STATE
	 $\hat{\mathbb{E}}[Y_{k} \otimes Y_{t} | X =x^{i}] = [y^{1}_{k} \otimes y^{1}_{t}, y^{2}_{k} \otimes y^{2}_{t}, \cdots, y^{n}_{k} \otimes y^{n}_{t}]^{\top} G$
	\STATE $S_{k,t} = |\det1(\hat{\mathbb{E}}[Y_{k} \otimes Y_{t}| X = x^{i}] )|$
	\ENDFOR
	 \STATE Compute  $D^{i}$ where $D^{i}[k,t] = -\log(\frac {S_{k,t}}{\sqrt{S_{k,k} \times S_{t,t}}})$
	 	\ENDFOR
\RETURN $ D_{L\times L} = \frac{1}{n}\sum_{i=1}^n D^{i} $
 		
 \end{algorithmic}
 \end{algorithm}

A significant amount of work has been done on learning latent tree models. Among the available approaches for latent tree learning, we use the information distance based algorithm
CLGrouping \cite{LTM} which has provable computational efficiency guarantees. These algorithms are based on a measure of statistical additive tree distance. For our conditional setting, we use the following form of the distance function:
\[\hat{d}_{kt}= \\ \frac{1}{n}\sum_{i=1}^n-\log\left(\frac {| \det1(\hat{\mathbb{E} }[Y_{k} \otimes Y_{t}| X=x^{i}])|}{\sqrt{S_{k,k}\cdot S_{t,t}}}\right),
 \] where $S_{k,k}:=| \det1(\hat{\mathbb{E} }[Y_{k} \otimes Y_{k}| X=x^{i}])|$, and similarly for $S_{t,t}$,  for observed nodes $k,t$ using $N$ samples. We employ the CL grouping to learn the tree structure from the estimated distances.

\section{ Learning CLTM Using Neural Networks}\label{sec:4}
Energy-based learning provides a unified framework for many probabilistic and non-probabilistic approaches to structured output tasks \cite{lossfunctions}, particularly for non-probabilistic training of graphical models and other structured models. Furthermore, the absence of the normalization condition allows for more flexibility in the design of learning machines. Most probabilistic models can be viewed as special types of energy-based models in which the energy function satisfies certain normalizability conditions, and in which the loss function, optimized by learning, has a particular form.
\subsection{Inference}

Consider observed variable X and output variable Y.
Define an energy function  $\mathcal{E}(X,Y)$ that is minimized  when $X$ and $Y$ are compatible. The most compatible $Y^{*}$ given an observed $X$ can be expressed as
\begin{equation*}
{ Y }^{ * }={\arg\min}_{ Y } \mathcal{E}(Y,X)
\end{equation*}
The energy function can be expressed as a factor graph, i.e. a sum of energy functions (node and edge potentials) that depend on input covariates x. Efficient inference procedures for factor graphs can be used to find the optimum configuration $Y^{*}$. In the below Eqn., we define the energy function which is used to model loss function.

 \begin{equation*}
 \mathcal{E}(x,z,\theta)= \sum_{k \in \mathcal{Z}}\phi_{k}(x, \theta)z_{k} + \sum_{(k,t) \in \mathcal{E}}\phi_{(k,t)}(x,\theta)z_{k}z_{t}
\end{equation*}

\subsection{Training Energy Based Models using Neural Networks}
Training an energy based model (EBM) consists of finding an energy function that produces the best Y for any X. The search for the best energy function is performed within a family of energy functions indexed by a parameter W. The architecture of the EBM is the internal structure of the parameterized energy function $\mathcal{E}(W,Y,X)$. In the case of neural networks the family of energy functions are the set of neural net architectures and weight values.

For a given neural network architecture, weights are learned by backpropagating the gradient through some loss function \cite{lossfunctions}. In the case of structures involving  latent variables h, we use negative marginal log-likelihood loss \eqref{loss} for training.
\begin{equation}
 \mathcal{L}=\mathbb{E}\left[\mathcal{E}(W,x,y,h)|y,x \right] - \mathbb{E}\left[\mathcal{E}(W,y,x,h)|x\right]
\label{loss}
\end{equation}

And the gradient is evaluated using below Eqn.
\begin{equation*}
\frac { \partial  \mathcal{L}} { \partial W }  = \mathbb{E}\left[ \frac { \partial  \mathcal{E}(W,y,x,h) }{ \partial W } |x,y \right] -  \mathbb{E}\left[ \frac { \partial \mathcal{E}(W,y,x,h) }{ \partial W } |x \right]
\end{equation*}

\section{Experiments}\label{sec:5}

In this section, we show experimental results of (a) classifying an image to multiple-object categories simultaneously and (b) identifying scenes from which images emerged. We use the non-iconic image data-set MS COCO \cite{MSCOCO} to evaluate our model. This data-set contains 83K training images with images labeled with 80 different object classes. The validation set contains 40K images. We use an independent classifier trained using 3 layer neural network (Indep. Classifier)  as a baseline, and compare precision-recall measures with our proposed conditional latent tree model.

\subsection*{Implementation}
We use our conditional latent tree model as a standalone layer on top of a neural network.
The layer takes as input a set of scores $\phi(x,W) \in \mathbb{R}^{n}$. These scores correspond to node potentials of the energy function. To avoid over-fitting we make edge potentials independent of input covariates. Using these potentials, our model outputs marginal probabilities of all the labels along with the MAP configuration. During learning, we use stochastic gradient descent and compute $ \frac{\partial  \mathcal{L}}{\partial \phi}$ , where $ \mathcal{L}$ is loss function defined in Eqn.\eqref{loss}. This derivative is then back propagated to the previous layers represented by $\phi(x;w)$. Using a mini-batch size of 250 and dropout, we train the model. We use the Viterbi message passing algorithm for exact inference on conditional latent tree model.

\begin{figure*}
\begin{center}
\psfrag{F-Measure}[l]{ \small{F-Measure}}
\psfrag{data1}[l]{ \small{CLTM}}
\psfrag{data2}[l]{ \small{3-layer NN}}
 \includegraphics[width=\linewidth]{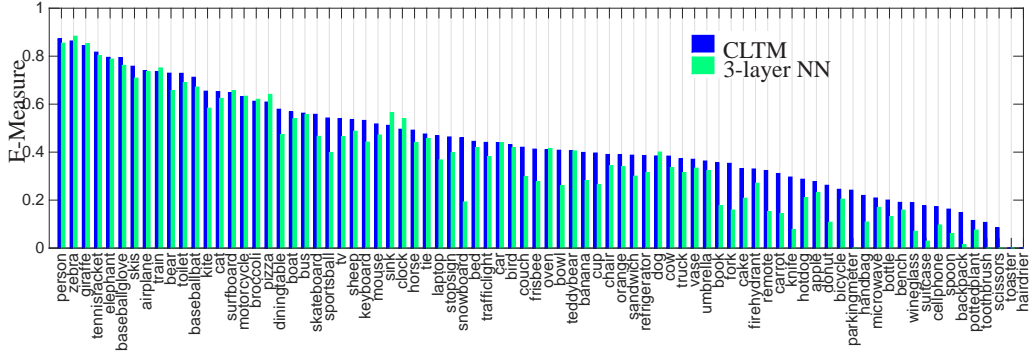}
\end{center}

   \caption{F-Measure comparison of individual classes}
\label{fig:f-measure}
\end{figure*}
\hspace{-1In}

\begin{figure*}
\begin{center}
\subfloat[]{
\centering{
\psfrag{Recall}[l]{ \tiny{Recall}}
\psfrag{Precision}[l]{ \tiny{Precision}}
\psfrag{CLTM}[l]{ \tiny{CLTM}}
\psfrag{Logistic Regression}[l]{ \tiny{3-layer NN}}
\psfrag{0}[l]{\tiny{0}}
\psfrag{0.1}[l]{\tiny{0.1}}
\psfrag{0.2}[l]{\tiny{0.2}}
\psfrag{0.3}[l]{\tiny{0.3}}
\psfrag{0.4}[l]{\tiny{0.4}}
\psfrag{0.5}[l]{\tiny{0.5}}
\psfrag{0.6}[l]{\tiny{0.6}}
\psfrag{0.7}[l]{\tiny{0.7}}
\psfrag{0.8}[l]{\tiny{0.8}}
\psfrag{0.9}[l]{\tiny{0.9}}
\psfrag{1}[l]{\tiny{1}}
  \includegraphics[width=0.3\linewidth]{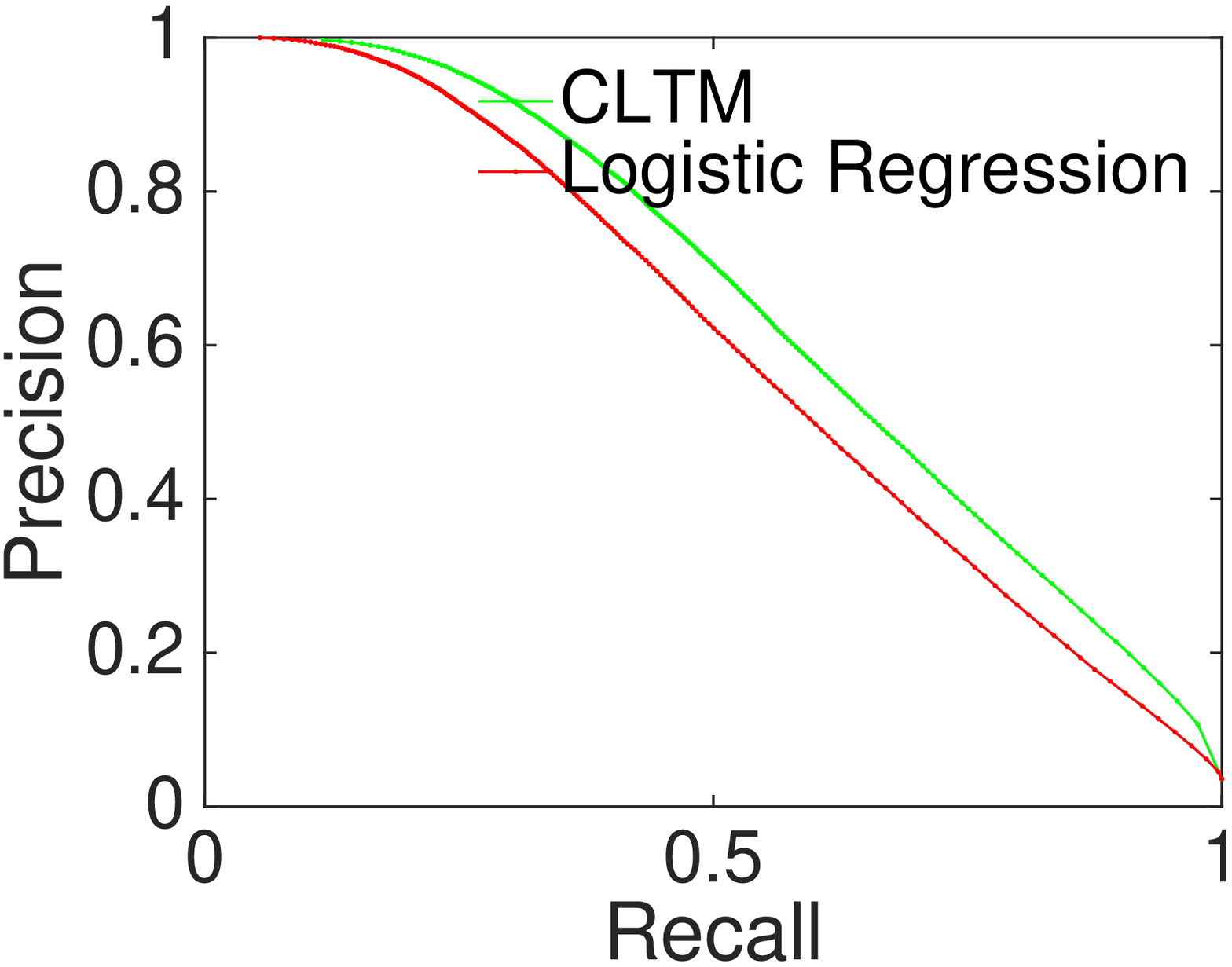}
}}
\subfloat[]{
\psfrag{Recall}[l]{ \tiny{Recall}}
\psfrag{Precision}[l]{ \tiny{Precision}}
\psfrag{CLTM}[l]{ \tiny{CLTM}}
\psfrag{Logistic Regression}[l]{ \tiny{3-layer NN}}
\psfrag{0}[l]{\tiny{0}}
\psfrag{0.1}[l]{\tiny{0.1}}
\psfrag{0.2}[l]{\tiny{0.2}}
\psfrag{0.3}[l]{\tiny{0.3}}
\psfrag{0.4}[l]{\tiny{0.4}}
\psfrag{0.5}[l]{\tiny{0.5}}
\psfrag{0.6}[l]{\tiny{0.6}}
\psfrag{0.7}[l]{\tiny{0.7}}
\psfrag{0.8}[l]{\tiny{0.8}}
\psfrag{0.9}[l]{\tiny{0.9}}
\psfrag{1}[l]{\tiny{1}}
\includegraphics[width=0.3\linewidth]{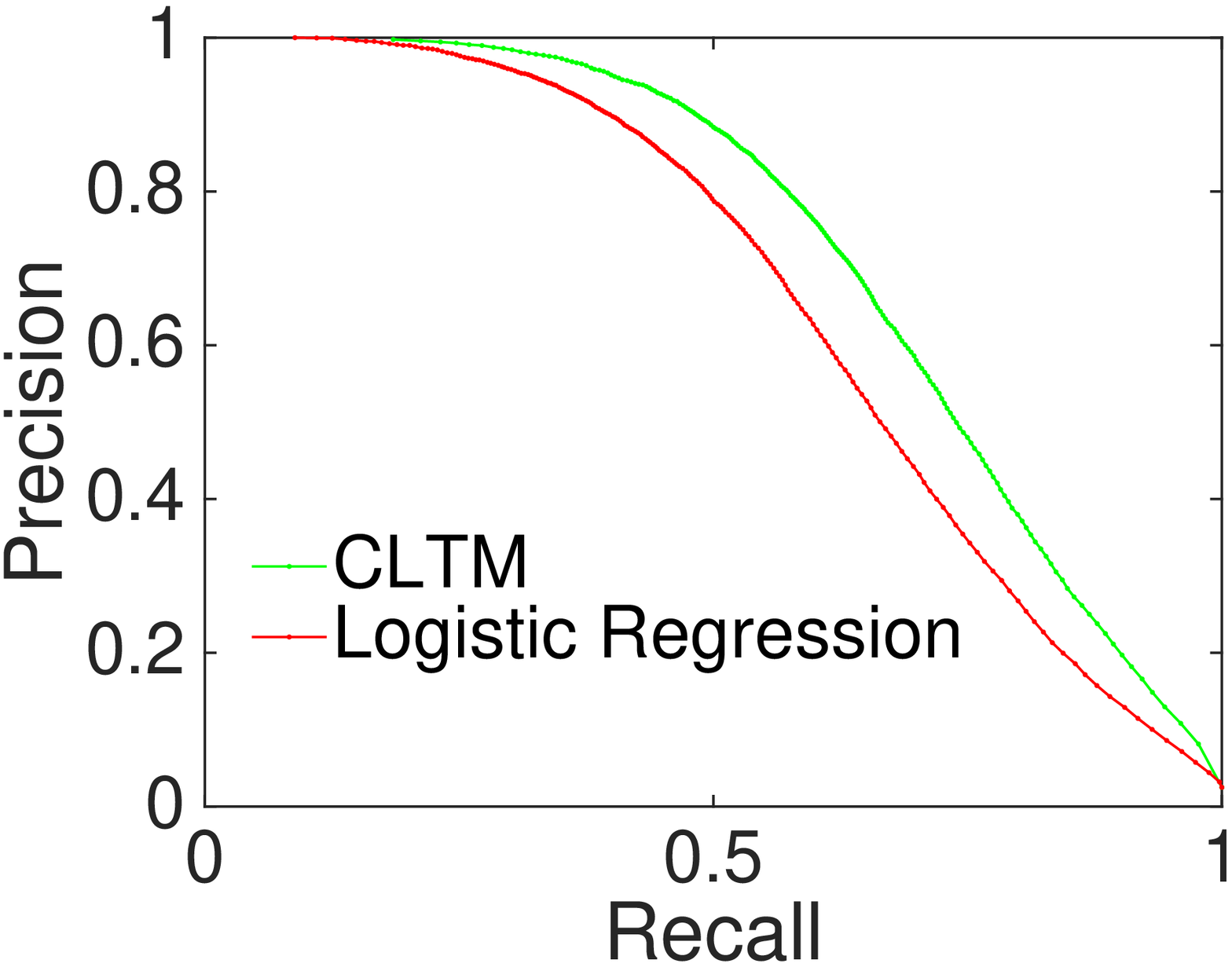}
}
\subfloat[]{
\psfrag{Recall}[l]{ \tiny{Recall}}
\psfrag{Precision}[l]{ \tiny{Precision}}
\psfrag{CLTM}[l]{ \tiny{CLTM}}
\psfrag{Logistic Regression}[l]{ \tiny{3-layer NN}}
\psfrag{0}[l]{\tiny{0}}
\psfrag{0.1}[l]{\tiny{0.1}}
\psfrag{0.2}[l]{\tiny{0.2}}
\psfrag{0.3}[l]{\tiny{0.3}}
\psfrag{0.4}[l]{\tiny{0.4}}
\psfrag{0.5}[l]{\tiny{0.5}}
\psfrag{0.6}[l]{\tiny{0.6}}
\psfrag{0.7}[l]{\tiny{0.7}}
\psfrag{0.8}[l]{\tiny{0.8}}
\psfrag{0.9}[l]{\tiny{0.9}}
\psfrag{1}[l]{\tiny{1}}
  \includegraphics[width=0.3\linewidth]{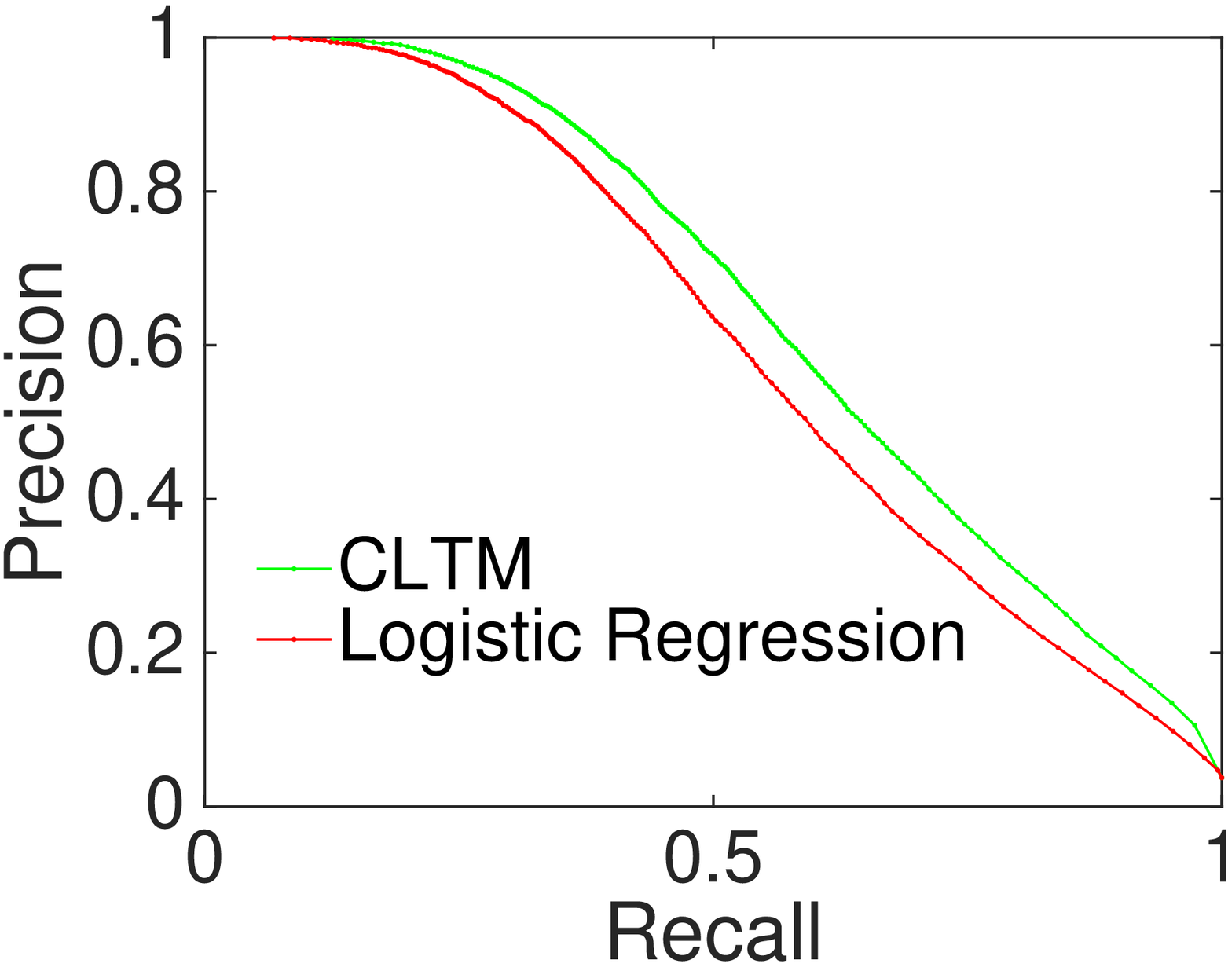}
}
\end{center}

\caption{Precision Recall Comparison: a) All the training images b) Subset of training images containing 2 object categories and c) Subset of training images containing 3 object categories. }
\label{fig:prcomparision}
\end{figure*}

\begin{figure}
\begin{center}
\subfloat{
\psfrag{keyboard}[l]{\small{}}
\psfrag{Recall}[l]{ \tiny{Recall}}
\psfrag{Precision}[l]{ \tiny{Precision}}
\psfrag{CLTM}[l]{ \tiny{CLTM}}
\psfrag{Logistic Regression}[l]{ \tiny{3-layer NN}}
\psfrag{0}[l]{\tiny{0}}
\psfrag{0.1}[l]{\tiny{0.1}}
\psfrag{0.2}[l]{\tiny{0.2}}
\psfrag{0.3}[l]{\tiny{0.3}}
\psfrag{0.4}[l]{\tiny{0.4}}
\psfrag{0.5}[l]{\tiny{0.5}}
\psfrag{0.6}[l]{\tiny{0.6}}
\psfrag{0.7}[l]{\tiny{0.7}}
\psfrag{0.8}[l]{\tiny{0.8}}
\psfrag{0.9}[l]{\tiny{0.9}}
\psfrag{1}[l]{\tiny{1}}
  \includegraphics[width=0.25\linewidth]{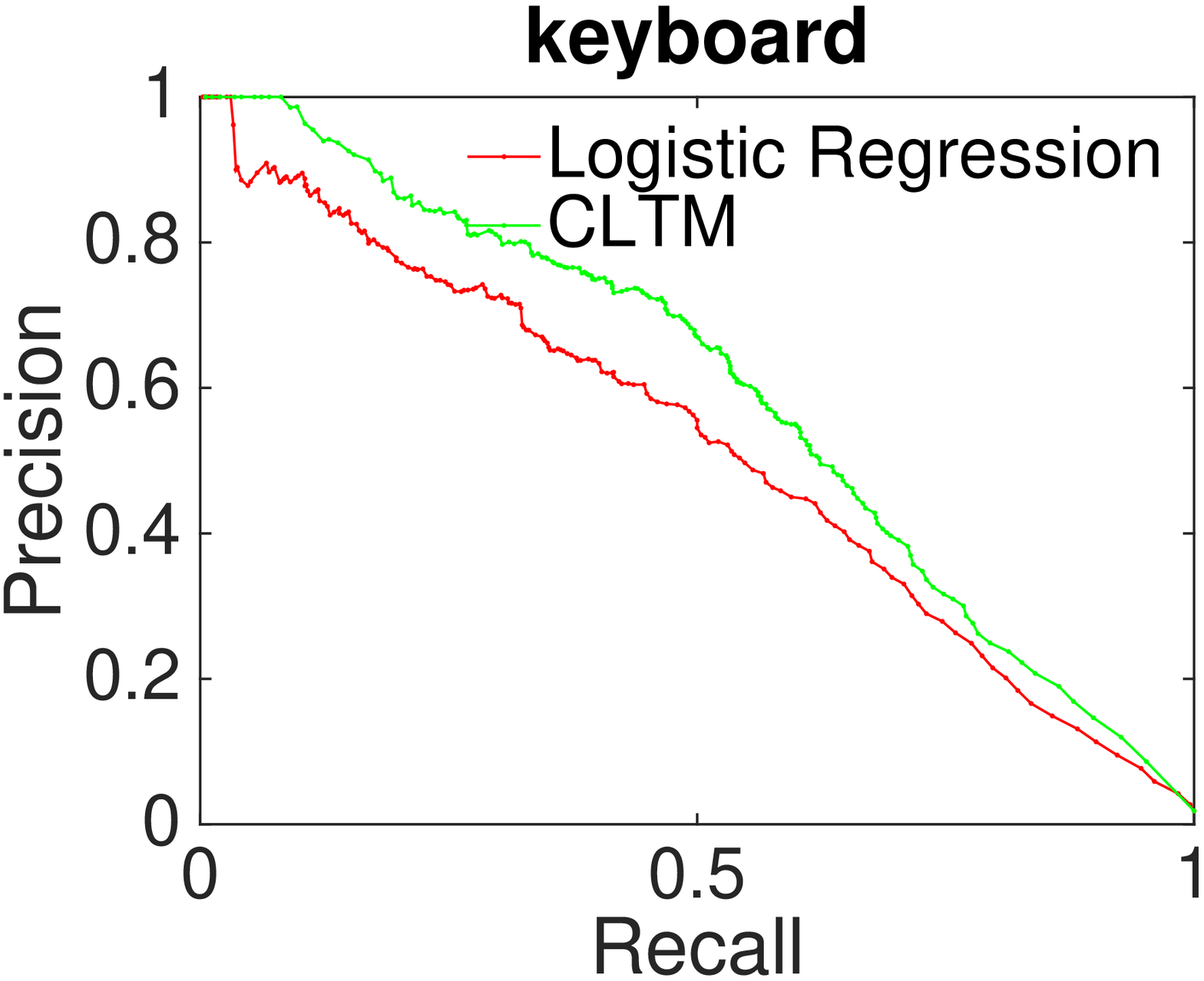}
}
\subfloat{
\psfrag{baseballglove}[l]{\small{}}
\psfrag{Recall}[l]{ \tiny{Recall}}
\psfrag{Precision}[l]{ \tiny{Precision}}
\psfrag{CLTM}[l]{ \tiny{CLTM}}
\psfrag{Logistic Regression}[l]{ \tiny{3-layer NN}}
\psfrag{0}[l]{\tiny{0}}
\psfrag{0.1}[l]{\tiny{0.1}}
\psfrag{0.2}[l]{\tiny{0.2}}
\psfrag{0.3}[l]{\tiny{0.3}}
\psfrag{0.4}[l]{\tiny{0.4}}
\psfrag{0.5}[l]{\tiny{0.5}}
\psfrag{0.6}[l]{\tiny{0.6}}
\psfrag{0.7}[l]{\tiny{0.7}}
\psfrag{0.8}[l]{\tiny{0.8}}
\psfrag{0.9}[l]{\tiny{0.9}}
\psfrag{1}[l]{\tiny{1}}
  \includegraphics[width=0.25\linewidth]{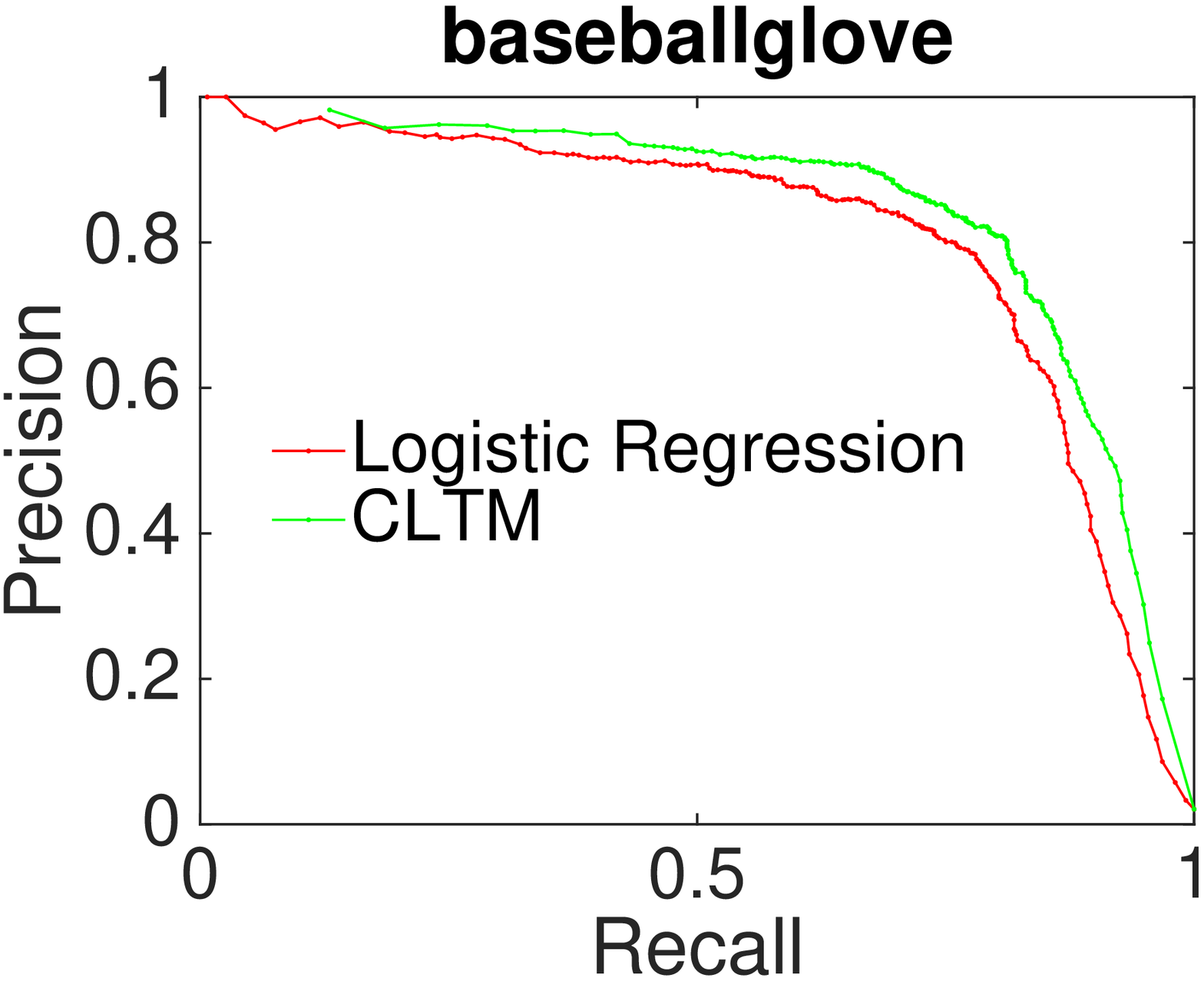}
}
\subfloat{
\psfrag{tennisracket}[l]{\small{}}
\psfrag{Recall}[l]{ \tiny{Recall}}
\psfrag{Precision}[l]{ \tiny{Precision}}
\psfrag{CLTM}[l]{ \tiny{CLTM}}
\psfrag{Logistic Regression}[l]{ \tiny{3-layer NN}}
\psfrag{0}[l]{\tiny{0}}
\psfrag{0.1}[l]{\tiny{0.1}}
\psfrag{0.2}[l]{\tiny{0.2}}
\psfrag{0.3}[l]{\tiny{0.3}}
\psfrag{0.4}[l]{\tiny{0.4}}
\psfrag{0.5}[l]{\tiny{0.5}}
\psfrag{0.6}[l]{\tiny{0.6}}
\psfrag{0.7}[l]{\tiny{0.7}}
\psfrag{0.8}[l]{\tiny{0.8}}
\psfrag{0.9}[l]{\tiny{0.9}}
\psfrag{1}[l]{\tiny{1}}
  \includegraphics[width=0.25\linewidth]{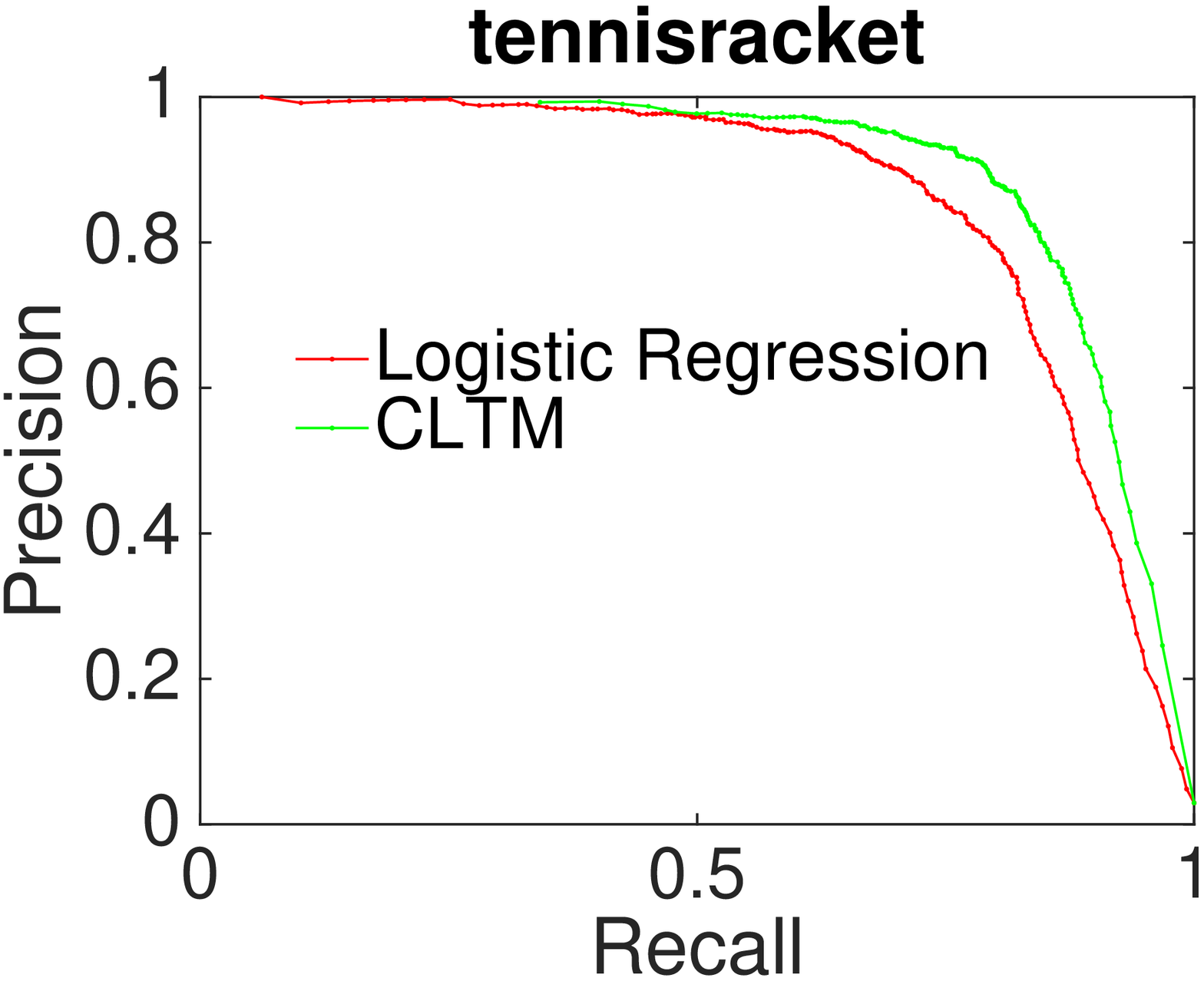}
}
\subfloat{
\psfrag{bed}[l]{\small{}}
\psfrag{Recall}[l]{ \tiny{Recall}}
\psfrag{Precision}[l]{ \tiny{Precision}}
\psfrag{CLTM}[l]{ \tiny{CLTM}}
\psfrag{Logistic Regression}[l]{ \tiny{3-layer NN}}
\psfrag{0}[l]{\tiny{0}}
\psfrag{0.1}[l]{\tiny{0.1}}
\psfrag{0.2}[l]{\tiny{0.2}}
\psfrag{0.3}[l]{\tiny{0.3}}
\psfrag{0.4}[l]{\tiny{0.4}}
\psfrag{0.5}[l]{\tiny{0.5}}
\psfrag{0.6}[l]{\tiny{0.6}}
\psfrag{0.7}[l]{\tiny{0.7}}
\psfrag{0.8}[l]{\tiny{0.8}}
\psfrag{0.9}[l]{\tiny{0.9}}
\psfrag{1}[l]{\tiny{1}}
  \includegraphics[width=0.25\linewidth]{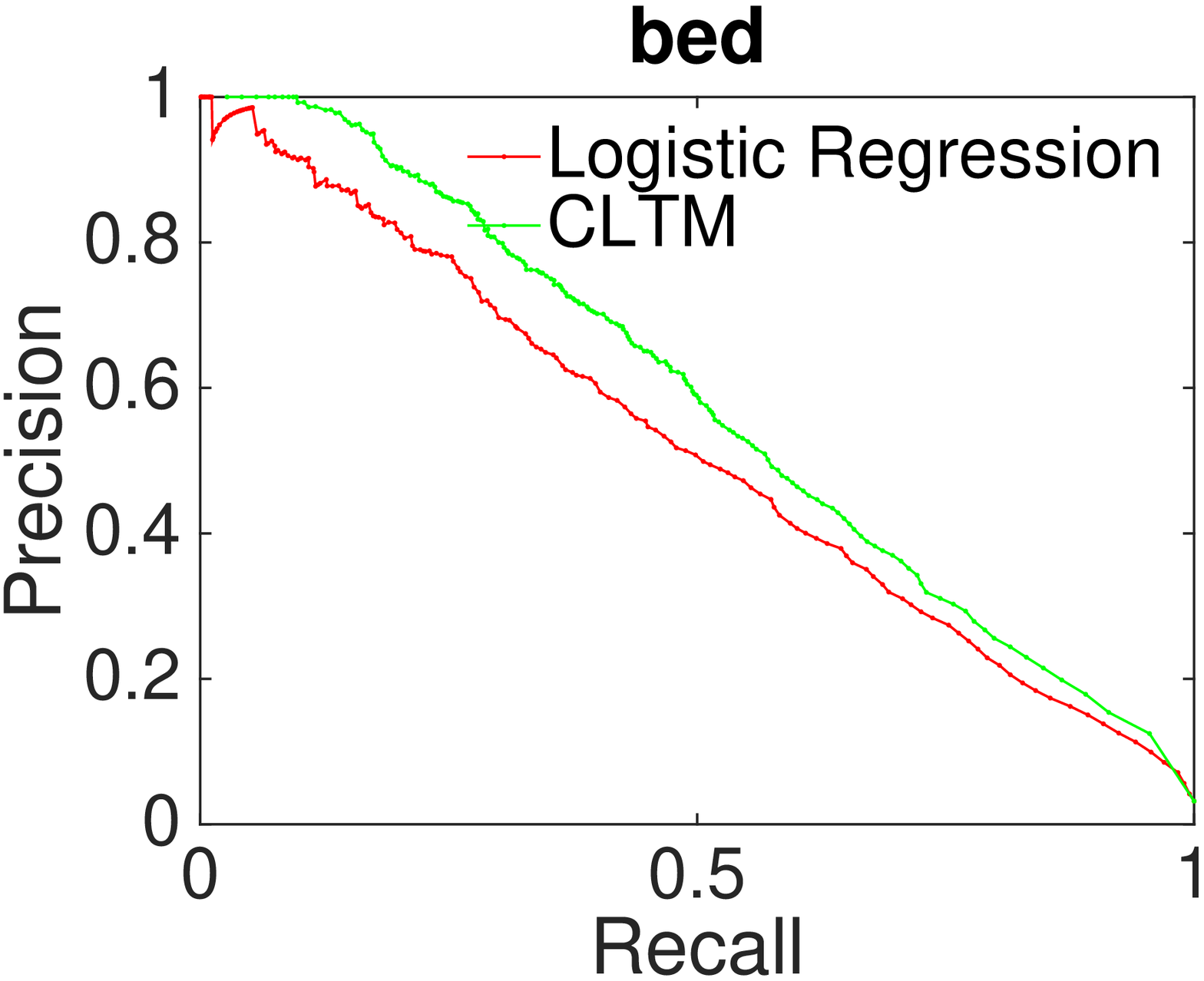}
}
\end{center}

\caption{Class-wise Precision-Recall for: a) Keyboard b) Baseball Glove c) Tennis Racket and d) Bed. }
\label{fig:classwise}
\end{figure}

\begin{table}[ht]
\caption{F-Measure  Comparison}

\centering
\begin{tabular}{|c c c c|}
\hline\hline
Model & Precision & Recall & F-Measure \\ [0.3ex]
\hline
1 Layer (Indep. Classifier) & 0.715 & 0.421 & 0.529 \\
1 Layer (CLTM) & 0.742 & 0.432 & 0.546 \\
2 Layer (Indep. Classifier) & 0.722 & 0.425 & 0.535 \\
2 Layer (CLTM) & 0.763 & 0.437 & 0.556 \\
3 Layer (Indep. Classifier) & 0.731 & 0.428 & 0.539 \\
3 Layer (CLTM)  & \textbf{0.769} & \textbf{0.449} & \textbf{0.567} \\
\hline
\end{tabular}
\label{table:nonlin}
\end{table}

\begin{figure}
\begin{center}

\includegraphics[width =0.9\linewidth]{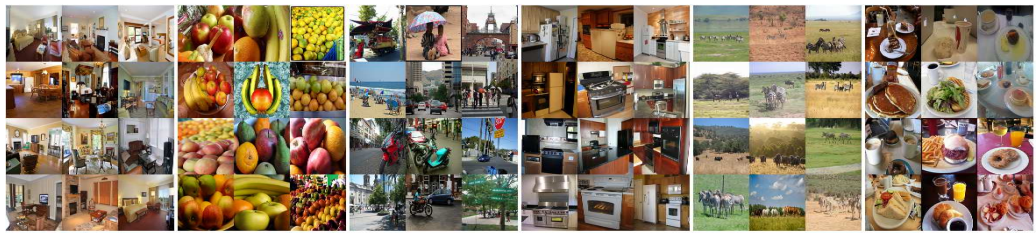}

\end{center}

\caption{Top 12 images producing the largest activation of node potentials for different latent nodes : (from left to right) $h17$  with neighborhood of objects appearing in living room ; $h5$ with neighborhood of objects belonging to class fruit ; $h3$ with neighborhood of objects appearing in outdoor scenes; $h4$ with neighborhood of objects appearing in kitchen ;$h9$ with neighborhood of objects appearing in forest; $h12$ with neighborhood of objects appearing on dining table. }
\label{fig:topkimages}

\end{figure}


\subsection{Structure Recovery}

We use 40k images randomly selected from the training set to learn the tree structure using the distance based method proposed in Section \ref{sec:3}. We have the recovered tree structure relating 80 different objects and 22 hidden nodes in \ref{appendix} Appendix. From the learned tree structure, we can see that hidden nodes take the role of dividing the tree according to the scene category. For instance, the nodes connected to hidden nodes $h19$, $h22$, $h9$ and $h17$  contain objects from the kitchen, bathroom, wild animals and living room respectively. Similarly, all the objects that appear in outdoor traffic scenes are clustered around the observed node car. Note that most training images contain fewer than 3 instances of different object categories.

\begin{figure*}[t]


\centering{\includegraphics[width = 0.8\linewidth]{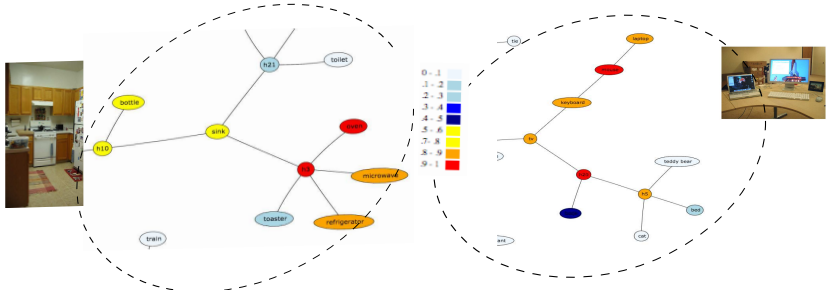}}

\caption{Figure showing heat map of marginal beliefs of nodes activated in different sub-trees for different images. }
\end{figure*}

\subsection{Classification Performance on MS COCO}

Table \ref{table:nonlin} shows the comparison of precision, recall and F-measure between 3 layer neural network independent classifier and Conditional Latent Tree Model trained using 1,2 and 3 layer feed forward neural networks respectively. For 3 layer neural network independent classifier, we use a threshold of 0.5 to make binary decisions for different object labels. For CLTM, we use the MAP configuration to make binary decisions. Note that CLTM improves F-measure significantly.  Fig.\ref{fig:f-measure} shows the
comparison of F-measure for each object category between baseline and CLTM trained using a 3 layer neural network.  Over-all the gain in F-measure using our model is 7-percent compared to 3 Layer neural network. Note that F-measure gain for indoor objects is more significant. For difficult objects like skateboard, keyboard, laptop, bowl, cup and wine-glass,  F-measure gain is 19-percent, 20-percent, 27-percent, 56-percent, 50-percent and 171-percent respectively. Fig. \ref{fig:prcomparision} shows the precision recall curves for a) entire test image set b) a subset of test images that contain 2 different object categories c) a subset of test images that contain 3 different object categories. We consider marginal probabilities of each observed class that our model produced to measure precision-recall curves for varying threshold values. Fig.\ref{fig:classwise} shows comparison of plots of precision-recall curves of a subset of object classes: tennis racket, bed, keyboard and baseball glove.

\subsection{Qualitative Analysis}

In this section, we investigate the class of images that triggered highest activation of node potentials for different latent nodes. Fig.~\ref{fig:topkimages} shows the top-12  images from test set that resulted in the highest activation of different latent nodes. It is observed that different latent nodes effectively capture different semantic information common to images containing neighboring object classes. For instance, the top-12 images of latent nodes $h9$ , $h12$, $h4$, $h21$, $h3$ and $h5$ resulted in a class of images appearing in scenes of forest, dining table, kitchen, living room, traffic and  belonging to fruit category.

\subsection{Scene Classification on MIT-Indoor Dataset}

The hidden nodes in CLTM model capture scene relevant information which can be used to perform scene classification tasks. In this section, we demonstrate scene classification capabilities of CLTM model. We use 529 images from MIT-Indoor data-set belonging to 4 different scenes: Kitchen, Bathroom, Living Room  and Bedroom. We perform k-means on outputs of CLTM model and 3 layer neural network independent classifier to cluster images. We then optimally match these clusters to scenes to evaluate misclassification rate. Note that we never trained our model using scene labels and we just use them for validating the performance. In our experiments, we use marginal  probabilities of observed and hidden nodes of CLTM , marginal probabilities of hidden nodes of CLTM  and probabilities of individual classes resulted from 3 layer neural network conditioned on input features. Table \ref{table:k-means} shows misclassification rates of different input features used for clustering.
With out the need of object presence knowledge, clustering on marginal probabilities of hidden nodes alone resulted in the least misclassification rate.

\begin{table}[ht]
\caption{Misclassification Rate}

\vspace{-0.15in}

\centering
\begin{tabular}{|c c c| }
\hline\hline
Model & k=4 & k=6  \\ [0.3ex]
\hline
Observed + Hidden & 0.326 & 0.242  \\
3 layer neural network & 0.390 & 0.301  \\
Hidden & \textbf{0.314} & \textbf{0.238}  \\
\hline
\end{tabular}
\label{table:k-means}
\end{table}

\section{Conclusion and Future Work} \label{sec:conclusion}
In conclusion, with the proposed structure recovery method we could recover the structure of latent tree. This tree has natural hierarchy of related objects placed according to their co-appearance in different scenes. We use neural networks of different architectures to train conditional latent tree models. We evaluate CLTM on MS COCO data-set and there is a significant gain in precision, recall and F-measure compared to 3 layer neural network independent classifier. Latent nodes captured different semantic information to distinguish high level class information  of images. Such an information is used for scene labeling task in an unsupervised manner.
In future, we aim to model both spatial and co-occurance knowledge and apply the model to object localisation tasks using CNN (like RCNN).

{\small
\bibliographystyle{ieee}
\bibliography{egbib}
}

\section*{Appendix}
\label{appendix}
\begin{figure}[ht]
\begin{center}

\includegraphics[width =\linewidth]{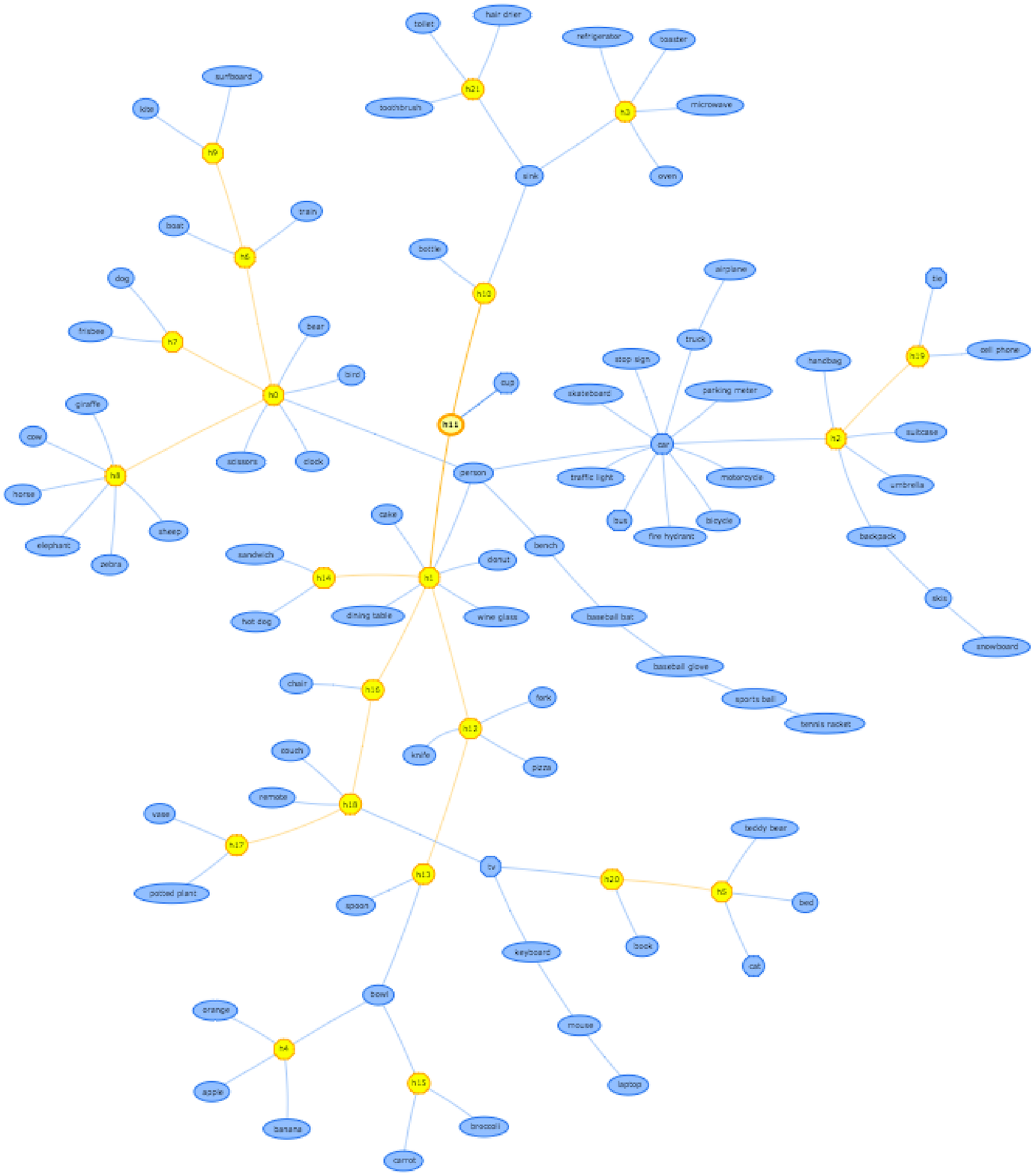}

\end{center}

\caption{Recovered tree structure}

\end{figure}

\end{document}